\documentclass[letterpaper, 10 pt, conference]{ieeeconf}  

\IEEEoverridecommandlockouts                              

\usepackage{graphics} 
\usepackage{epsfig} 
\usepackage{times} 
\usepackage{amsmath} 
\usepackage{amssymb}  
\usepackage{mathrsfs}
\usepackage{multirow} 

\def\ie{{\em i.e.}}
\def\eg{{\em e.g.}}
\def\etal{{\em et al.}}

\title{2D LiDAR Map Prediction via Estimating Motion Flow with GRU}

\author{Yafei Song$^{12}$, Yonghong Tian$^{1}$, Gang Wang$^{2}$ and Mingyang Li$^{2}$
\thanks{$^{1}$Yafei Song and Yonghong Tian are with the National Engineering Laboratory for Video Technology, School of Electronics Engineering and Computer Science, Peking University, Beijing 100871, China. {\tt\small \{songyf, yhtian\}@pku.edu.cn}}%
\thanks{$^{2}$Yafei Song, Gang Wang and Mingyang Li are with the AI Labs, Alibaba Group, Hangzhou 311121, China. {\tt\small \{huaizhang.syf, wg134231, na.lmy\}@alibaba-inc.com}}%
}

\begin{document}

\maketitle
\thispagestyle{empty}
\pagestyle{empty}

\begin{abstract}
It is a significant problem to predict the 2D LiDAR map at next moment for robotics navigation and path-planning. To tackle this problem, we resort to the motion flow between adjacent maps, as motion flow is a powerful tool to process and analyze the dynamic data, which is named optical flow in video processing. However, unlike video, which contains abundant visual features in each frame, a 2D LiDAR map lacks distinctive local features. To alleviate this challenge, we propose to estimate the motion flow based on deep neural networks inspired by its powerful representation learning ability in estimating the optical flow of the video. To this end, we design a recurrent neural network based on gated recurrent unit, which is named LiDAR-FlowNet. As a recurrent neural network can encode the temporal dynamic information, our LiDAR-FlowNet can estimate motion flow between the current map and the unknown next map only from the current frame and previous frames. A self-supervised strategy is further designed to train the LiDAR-FlowNet model effectively, while no training data need to be manually annotated. With the estimated motion flow, it is straightforward to predict the 2D LiDAR map at the next moment. Experimental results verify the effectiveness of our LiDAR-FlowNet as well as the proposed training strategy. The results of the predicted LiDAR map also show the advantages of our motion flow based method.
\end{abstract}

\section{Introduction}

\begin{figure}
  \includegraphics[width=\linewidth]{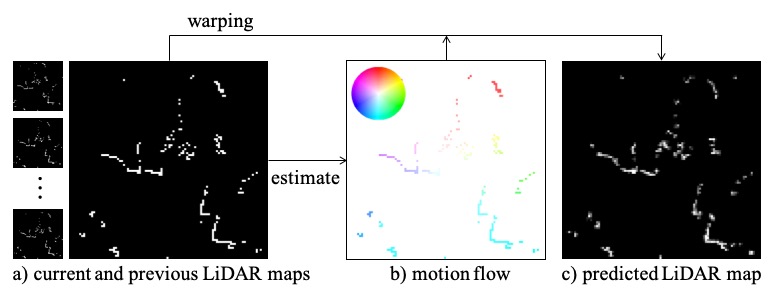}
  \caption{In this paper, we predict the LiDAR map at next moment via estimating the motion flow only from the current and previous maps.}
  \label{fig:teaser}
\end{figure}

2D light detection and ranging (LiDAR) sensor is widely equipped on autonomous vehicles and various types of robotics due to its robustness and lower price. A 2D LiDAR sensor can detect the obstacles in the scene and measure their distance. However, as shown in Fig.~\ref{fig:teaser}, the scan data are binary when they are transformed into a 2D map, which contains much less information than some other sensors such as 3D LiDAR or image/video sensors. Due to this limitation, it is very challenging to perform some high-level perception tasks only with 2D LiDAR, such as to predict the future map in a dynamic scene, which is a significant problem for robotics navigation and path-planning. 

To predict the dynamic map, previous methods can be divided into three classes, state based methods,  direct methods, and motion flow based methods. State based methods are also closely related to moving object tracking problem. With the help of odometry information, these methods classify the cells in a LiDAR map into two classes, moving and static. Then, the next map can be predicted by estimating the state of moving objects. To this end, traditional methods~\cite{zhao1998qualitative, arras2007using, petrovskaya2009model, vu2007online, yang2011simultaneous, wang2015model} usually divide the pipeline into several steps, including to detect separate objects, associate measurements with tracked objects, estimate the state, and predict the next state of each tracked object. Then, the next map can be predicted. These methods include several hand-designed processes along with some parameters. The parameters should be manually tuned, which limits their applicability. 

For direct methods, Ondr{\'{u}}{\v{s}}ka \etal~\cite{ondruska2016deep, ondruska2016end, DequaireIJRR2018} proposed the DeepTracking method based on gated recurrent unit (GRU) \cite{cho2014learning} to directly predict the occupancy map. DeepTracking formulates the problem as a segmentation problem. For each cell in the map, DeepTracking predicts the probability whether it will be occupied or not at the next moment. In contrast, our method predicts the map via estimating motion flow. As motion flow encodes the velocity of each cell, we can obtain the predicted map as well as the motion information of each cell.

As is well-known, optical flow~\cite{gibson1950perception} encodes the correspondence relationship between adjacent frames, which is a powerful tool in processing and understanding the contents in a video. However, as shown in Fig.~\ref{fig:teaser}, a 2D LiDAR map only contains sparse occupied points of the scene, which leads to inadequate context information and unrepresentative local features. Therefore, it is a challenging problem to estimate the optical flow of 2D LiDAR. Moreover, as the word \textit{optical} usually refers to visual sensors, we use \textit{motion flow} instead in this paper. To estimate motion flow, previous works resorted to Bayesian occupancy filter~\cite{chen2006dynamic, gindele2009bayesian}, hidden Markov model~\cite{meyer2012occupancy} and recurrent flow network~\cite{Choi2016}. The presentation ability of these models, however, is usually too weak to cover this problem, which leads to unsatisfactory performance. Moreover, in these methods, there are many parameters to be empirically adjusted, which limits their application.

As shown in Fig.~\ref{fig:teaser}, we also predict the next map via motion flow. To alleviate the challenge, we resort to deep learning based method, which is a data-driven method and can automatically learn representative features from the data. Moreover, as the next frame is unknown, we can only exploit the current and previous frames. To this end, we design a recurrent network based on gated recurrent unit, which is named LiDAR-FlowNet. With the help of GRU, the spatiotemporal motion information can be effectively captured and encoded. Our LiDAR-FlowNet can simultaneously estimate backward and forward motion flow between the current map and the unknown next map. A self-supervised strategy is further designed to train the LiDAR-FlowNet model. No training data need to be annotated with this strategy. To alleviate the challenge in processing the binary maps, we propose to filter the maps using Gaussian kernels, which makes the training process more effective. With the backward motion flow, we can easily predict the next dynamic map via warping the current map. Experimental results verify the effectiveness of our LiDAR-FlowNet as well as the proposed training strategy. The results of the predicted map also show the advantages of our motion flow based method.

Our contributions mainly lie in three aspects:
\begin{enumerate}
\item We propose a recurrent neural network named LiDAR-FlowNet which can estimate the motion flow between current 2D LiDAR map and unknown next map only based on the current map and previous maps.
\item We design a self-supervised strategy along with Gaussian filter to train the LiDAR-FlowNet effectively. No training data need to be annotated with this strategy.
\item With the estimated motion flow, we can predict the dynamic LiDAR map at the next moment. The experimental results verify the usefulness and effectiveness of motion flow as well as our proposed LiDAR-FlowNet. 
\end{enumerate}

\section{Related Work}

To predict the dynamic 2D LiDAR map, previous methods can be divided into three classes, state based methods, direct methods, and motion flow based methods. We will briefly review these methods as follows. 

\textbf{State based methods}. State based methods are also closely related to moving object tracking problem. These methods usually exploit the odometry information, and then the dynamic cells are only caused by the moving objects. These methods first classify the cells into two classes, moving and static. The next map can be predicted by estimating the state of moving objects. The whole process can be divided into several steps, including detection of separate objects, association of measurements with tracked objects, estimation of the state, and prediction of the next state of each tracked object. Then, the next map can be predicted. Following this framework, some methods~\cite{vu2007online, yang2011simultaneous, wang2015model} aim at tackling the general motion of arbitrary types of objects. However, this problem is too complicated for traditional methods, and the performance is unsatisfying. On the other hand, some other methods~\cite{zhao1998qualitative, arras2007using, petrovskaya2009model} only consider some presupposed motion patterns for the tracked objects. These methods can only improve the performance for the right motion patterns.
Moreover, these methods include several hand-designed processes along with some parameters. The parameters should be manually tuned, which limits the applicability of these methods. 

\textbf{Direct methods}. For direct methods, Ondr{\'{u}}{\v{s}}ka \etal~\cite{ondruska2016deep, ondruska2016end, DequaireIJRR2018} also resorted to deep learning based method. They formulated the problem as a segmentation problem, which is to segment the occupied cells from the map. To this end, they designed a recurrent neural network based on the gated recurrent unit (GRU). For each cell, the deep model predicts the probability whether it will be occupied or not at the next moment. To train the model, they took the occupancy map at the next moment as ground truth and minimized a cross entropy loss. The network they used actually is similar with ours. However, our method predicts the map via estimating motion flow and the training strategy is also different. As motion flow encodes the velocity of each cell, we can obtain the predicted map as well as the motion information of each cell. The estimated motion flow could also be a powerful tool to perform other tasks.

\textbf{Motion flow based methods}. To estimate the motion flow of 2D LiDAR, Chen~\etal~\cite{chen2006dynamic} proposed Bayesian occupancy filter. Gindele~\etal~\cite{gindele2009bayesian} further extended this method by incorporating prior knowledge. Choi~\etal~\cite{Choi2016} proposed a recurrent flow network. Note that, the proposed network is different from the conception in the deep learning community. The proposed network mainly consists of a context layer, which is to encode the velocity of each cell in the 2D LiDAR scan map. The velocity also can be regarded as the motion flow. All these methods model the variation of each cell and its local neighbors and then estimate the motion flow. However, the presentation ability of these shallow models usually is too weak to encode the complex problem, which leads to unsatisfying performance. Moreover, in these methods, there are many parameters to be empirically adjusted, which limits their application.


\begin{figure*}
  \includegraphics[width=17cm]{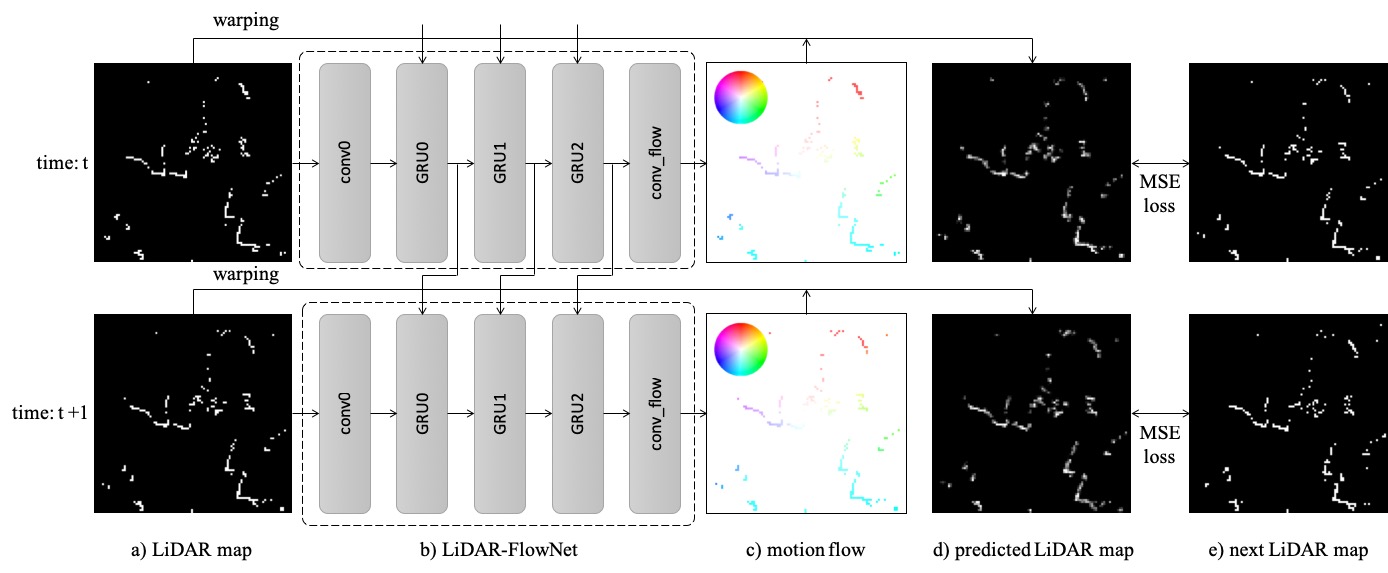}
  \caption{The pipeline of our method. At each moment, the occupancy and visibility maps along with the hidden states are fed into the LiDAR-FlowNet. The LiDAR-FlowNet can predict the backward motion flow. Then the next frame can be estimated via warping the current frame. The mean squared error between the estimated next frame and the real next frame can be used as the loss function to train the LiDAR-FlowNet model.}
  \label{fig:pipeline}
\end{figure*}

\section{Overview: Map Prediction Using Motion Flow}

\begin{figure}
\center	
  \includegraphics[width=5cm]{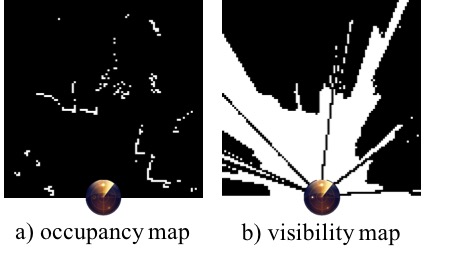}
  \caption{A pair of binary maps transformed from a LiDAR scan, including the occupancy map a) and the visibility map b).}
  \label{fig:occ:vis}
\end{figure}

In this section, we briefly introduce the pipeline of our method. For each 2D LiDAR scan, we place the LiDAR at the bottom-middle of the map and convert the scan data from a vector to two binary maps, occupancy map $\mathcal{O}$ and visibility map $\mathcal{V}$. A pair of example maps are demonstrated in Fig.~\ref{fig:occ:vis}. In an occupancy map, a cell $\mathcal{O}_i = 1$ if it is occupied, and $\mathcal{O}_i = 0$ if it is free. In a visibility map, a cell $\mathcal{V}_i = 1$ if it is visible, and $\mathcal{V}_i = 0$ if it is occluded by occupied cells. In this form, all the information contained in the LiDAR scan can be encoded in the 2D maps, which can be efficiently processed by deep learning base method. Note that, we limit the FoV of the LiDAR to $180^{\circ}$.

We denote the backward motion flow between current frame $\mathcal{O}^t$ and next frame $\mathcal{O}^{t+1}$ as $\mathcal{B}^t$, where $\mathcal{O}^{t+1}_i = \mathcal{O}^t_{i+\mathcal{B}^t_i}$. Similarly, the forward motion flow is denoted as $\mathcal{F}^t$, where $\mathcal{O}^{t}_i = \mathcal{O}^{t+1}_{i+\mathcal{F}^t_i}$. Moreover, $\mathcal{B}_i$ and $\mathcal{F}_i$ are not always integer, if not, we use bilinear sampling to perform the warping process. According to the definition, we can see that the occupancy map at the next moment can be easily calculated if we have the backward motion flow. The key problem is transformed into estimating the motion flow.

As shown in Fig.~\ref{fig:pipeline}, we demonstrate the pipeline of our method. At each moment, we feed the occupancy and visibility maps along with the hidden states of GRUs at the previous moment into the LiDAR-FlowNet. Moreover, the hidden states could encode the motion information of each cell. With the LiDAR-FlowNet, we can predict the backward motion flow of the current frame. Then the next frame can be estimated via warping the current frame according to the motion flow. The mean squared error between the estimated next frame and the real next frame can be used as the loss function to train the LiDAR-FlowNet model. Note that, we simultaneously estimate the forward motion flow, which is not in the pipeline for simplicity.

\section{Estimate Motion Flow of 2D Lidar}

In this section, we detailedly introduce each step of our method, including the LiDAR-FlowNet, the self-supervised training strategy, and the Gaussian filter which is used to facilitate the training process. 

\subsection{LiDAR-FlowNet With GRU}

To estimate the motion flow using deep neural networks, previous methods~\cite{Dosovitskiy2015, Ilg2017} usually adopt feed forward neural networks, \ie~convolutional neural networks. These methods need a pair of frames as input and estimate the motion flow between them via implicitly or explicitly feature matching. However, for our problem, we aim to predict the next frame, which is unknown. Therefore, feed-forward neural networks are not suitable for our problem. On the other hand, recurrent neural networks can encode and exploit the dynamic history information, which can be used to estimate motion flow. Inspired by DeepTracking~\cite{DequaireIJRR2018}, we believe that recurrent neural networks could estimate the motion flow between the current frame and the unknown next frame only from the current frame and the previous frames.

To this end, we design a recurrent neural network named LiDAR-FlowNet, which is demonstrated in Fig.~\ref{fig:pipeline}. Our LiDAR-FlowNet consists of six layers. The first layer is the input layer, which reads a pair of occupancy and visibility maps and feeds them into the network. The second layer is a convolution layer, which aims to extract local features from the input. The third, fourth and fifth layers are gated recurrent layers~\cite{cho2014learning}, which actually is a convolutional gated recurrent unit. For each GRU, the output $y^t$ at time $t$ can be calculated from its input $x^t$ at the current time $t$ and output $y^{t-1}$ at the previous time $t-1$ as 
\begin{equation}
\begin{split}
f^t & = \sigma ( W_{xf} \circledast x^t + W_{yf} \circledast y^{t-1} ), \\
r^t & = \sigma ( W_{xr} \circledast x^t + W_{yr} \circledast y^{t-1} ), \\
\overline{y}^t & = \tanh ( W_{xy} \circledast x^t + r^t \cdot W_{yy} \circledast y^{t-1} ), \\
y^t & =  f^{t} \cdot y^{t-1} + (1 - f^{t}) \cdot \overline{y}^t,
\end{split}
\end{equation}
where $f^t$ is the update gate, $r^t$ is the reset gate, $\circledast$ denotes convolution operation, and $\cdot$ denotes dot product operation. The details of GRU can be found in \cite{cho2014learning}. With the help of GRU, the model can encode temporal information of each cell, which can facilitate the motion flow estimation.

The last layer is the output layer, which is also a convolution layer and calculates the forward and backward motion flow. As the motion flow of each cell consists of the horizontal and vertical offset, the output has four channels. The parameters of each layer are listed in Table~\ref{tab:net:structure}. Besides the standard convolution operation, we use dilation~\cite{Chen2016} to enlarge the perception field while maintaining the computation unchanged. With the help of dilation, the model could encode fast moving objects in the map.

\begin{table}
\caption{The structure of our LiDAR-FlowNet, where f is short for filter size, s for stride, d for dilation, p for padding.}
\label{tab:net:structure}
\begin{center}
\begin{tabular}{ @{\hspace{2mm}} c | @{\hspace{2mm}} c | @{\hspace{2mm}} c @{\hspace{2mm}} | c @{\hspace{2mm}}}
\hline \hline
Layer name & Type & Parameters & Output size \\
\hline
input & - & - & $2 \times 100 \times 100$ \\
\hline
\multirow{2}{*}{conv0} & \multirow{2}{*}{Convolution} & f: $3\times3$, s: 1, & \multirow{2}{*}{$16 \times 100 \times 100$} \\
 & & d: 1, p:1 &  \\
\hline
\multirow{2}{*}{gru0} & \multirow{2}{*}{GRU} & f: $3\times3$, s: 1, & \multirow{2}{*}{$16 \times 100 \times 100$} \\
 &  & d: 1, p:1 &  \\
\hline
\multirow{2}{*}{gru1} & \multirow{2}{*}{GRU} & f: $3\times3$, s: 1, & \multirow{2}{*}{$16 \times 100 \times 100$} \\
 &  & d: 2, p:2 &  \\
\hline
\multirow{2}{*}{gru2} & \multirow{2}{*}{GRU} & f: $3\times3$, s: 1, & \multirow{2}{*}{$16 \times 100 \times 100$} \\
 &  & d: 4, p:4 &  \\
\hline
\multirow{2}{*}{conv\_flow} & \multirow{2}{*}{Convolution} & f: $3\times3$, s: 1, & \multirow{2}{*}{$4 \times 100 \times 100$} \\
 & & d: 1, p:1 &  \\
\hline \hline
\end{tabular}
\end{center}
\end{table}

\subsection{Self-supervised Training Strategy}

The ground truth motion flow is difficult to be manually annotated. To this end, synthetic datasets, \eg~FlyingChairs~\cite{Dosovitskiy2015}, are generated and popular. However, the synthetic data are not the same as real data. On the other hand, some self-supervised methods~\cite{ZhouECCV2016, LiuICCV2017, ren2017unsupervised, Zhong2017, Godard2017} are proposed to avoid data synthesis and annotation. Inspired by these methods, we propose a similar strategy to train our LiDAR-FlowNet model. As shown in Fig.~\ref{fig:pipeline}, at time $t$, the input occupancy map $\mathcal{O}^t$ can be warped as $\mathcal{\hat{O}}^{t+1}$ according to the estimated backward flow. The warped occupancy map $\mathcal{\hat{O}}^{t+1}$ should be similar to the occupancy map at the next time $\mathcal{O}^{t+1}$. To evaluate the similarity between them, we use the mean squared error. Then the LiDAR-FlowNet model can be trained via minimizing this error. Moreover, the occupancy map at the next time $\mathcal{O}^{t+1}$ also can be warped as $\mathcal{\hat{O}}^{t}$ according to the forward motion flow. As forward motion flow and backward motion flow are relevant, we can formulate them as a multi-task problem and simultaneously estimate them. Usually, if we perform several relevant tasks using a multi-task model, the performance of all tasks will be improved. Then the loss function can be defined as 
\begin{equation}
\begin{split}
\mathcal{L} & = \mathtt{MSE} (\mathcal{O}^{t},\mathcal{\hat{O}}^{t}) + \mathtt{MSE}(\mathcal{O}^{t+1},\mathcal{\hat{O}}^{t+1}) \\
& = \mathtt{MSE} (\mathcal{O}^{t}, \mathtt{W}(\mathcal{O}^{t+1}, \mathcal{F}^t)) + \mathtt{MSE}(\mathcal{O}^{t+1}, \mathtt{W}(\mathcal{O}^{t}, \mathcal{B}^t)),
\end{split}
\label{eq:loss}
\end{equation}
where $\mathtt{MSE}()$ is the mean squared error function and $\mathtt{W}()$ is the warping function.

Moreover, to train the model, the warping step should be differentiable. This problem has been thoroughly investigated in spatial transform network \cite{Jaderberg2015}. More specifically, we use bi-linear sampling in the forward calculation process.

\subsection{Facilitate the Training Process Using Gaussian Filter}

\begin{figure}
\center	
  \includegraphics[width=8cm]{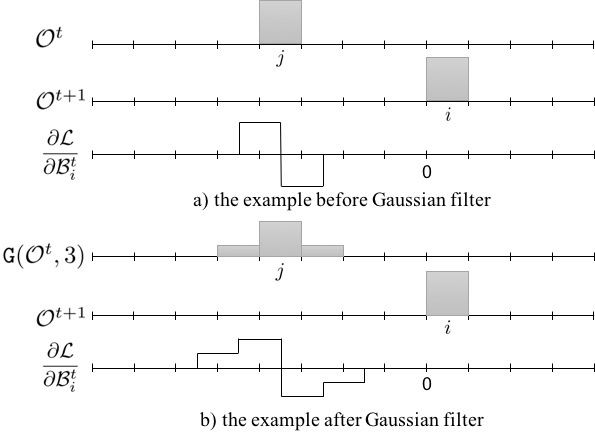}
  \caption{A large proportion of cells are zeros in the occupancy map, which leads to zero gradients for a wide range of estimated motion flow, as the example shown in a). If the input occupancy map is filtered by a Gaussian filter, we can enlarge the non-zero gradient range, as shown in b).}
  \label{fig:problem:gaussian}
\end{figure}

The proposed LiDAR-FlowNet can be trained via minimizing the loss function \eqref{eq:loss}. However, as the occupancy map is very different from the rgb image, the training process is not effective. The problem is that as shown in Fig.~\ref{fig:occ:vis}~a), a large proportion of cells are zero in the occupancy map, and the most gradients will be zero. To illustrate this problem, we take a pair of simple one-dimension occupancy maps as an example. As shown in Fig~\ref{fig:problem:gaussian}~a), the gradient $\frac{\partial \mathcal{L}}{\partial \mathcal{B}^t_i}$ is non-zero only when the motion flow $\mathcal{B}^t_i \in (j-i-1, j-i+1)$. This means that the gradient will be zero in most instances, which will lead to ineffective training. 

To alleviate this problem, we propose to filter the occupancy map using the Gaussian kernel before warping. As shown in Fig~\ref{fig:problem:gaussian}~b), the Gaussian filter will enlarge the range of non-zero gradient. Then the loss function will be 
\begin{equation}
\begin{split}
\mathcal{L} = & \mathtt{MSE} (\mathcal{O}^{t}, \mathtt{W}(\mathtt{G}(\mathcal{O}^{t+1}, \mathtt{f}), \mathcal{F}^t)) + \\
& \mathtt{MSE} (\mathcal{O}^{t+1}, \mathtt{W}(\mathtt{G}(\mathcal{O}^{t}, \mathtt{f}), \mathcal{B}^t)).
\end{split}
\label{eq:loss:gaussian}
\end{equation}
where $\mathtt{G}()$ is the Gaussian filter operation and $\mathtt{f}$ is the filter size. We first set a large filter size $\mathtt{f}$, and gradually decrease it until it is $1$ along with the training process. When $\mathtt{f}=1$, the loss function \eqref{eq:loss:gaussian} is the same as \eqref{eq:loss}.

\section{Experiments}

In this section, we first explain the experimental details of our LiDAR-FlowNet model. Then, we present the results of LiDAR map prediction to verify the effectiveness of our method as well as the LiDAR-FlowNet. The results of motion flow estimation are also presented to demonstrate the ability of our LiDAR-FlowNet. 

\subsection{Experimental Details of LiDAR-FlowNet}

The robotics platform may be static or dynamic relative to the scene. Thus, the data also can be divided into two classes accordingly. It is obvious that the dynamic scenario is more difficult than the static scenario. We set up a robotic platform to collect these two types of data in our indoor office. The robotic platform is equipped with a 2D LiDAR sensor, \ie~SICK TIM561-2050101. The LiDAR scans the scene at $15$ fps. For each scenario, we collect about $40$ minutes data, \ie~$36,000$ frames, for training and $8$ minutes data, \ie~$7,200$ frames, for validation. For the static scenario, as the robotic platform is static relative to the scene, the moving objects are the walking people. For the dynamic scenario, all the things are moving relative to the robotic platform. Each scan is transformed into an occupancy map and a visibility map. The map size is $100 \times 100$, and the resolution of each cell is $10$ centimeter.

To train the LiDAR-FlowNet model, we divide the training data into $1,800$ sequences, and each sequence consists of $20$ frames. For each sequence, the first $10$ frames are fed into the LiDAR-FlowNet to initialize the hidden state. From the eleventh frame, we take the mean squared error \eqref{eq:loss:gaussian} as the loss, and minimize this loss to update the model. Specifically, for RNN like LiDAR-FlowNet, a sequence is a training sample. In our experiment, we set the batch size as $32$, the initial learning rate as $0.01$. The learning rate is divided by $2$ after every $25$ epoch. The final model is obtained after $200$ epochs training. For the Gaussian filter, we set the initial filter size as $9$, and reduce the filter size by $2$ after every $25$ epoch until it equals to $1$.

To implement the LiDAR-FlowNet, we resort to the deep learning framework pytorch~\cite{PyTorchNIPS2017}. We use an NVIDIA GPU P100 to train the model, and it takes about $10$ minutes to perform one epoch.

\subsection{LiDAR Map Prediction Results}

With the LiDAR-FlowNet model, we can predict the motion flow of each cell in the scan map of current frame. Then the scan map of next frame can be obtained via warping the current scan map. For each cell in the predicted map, it will be $1$ when it is more than a threshold, otherwise $0$. We empirically set the threshold parameter as $0.4$. To quantitatively evaluate the results, we use the metric $F_1$ score and precision-recall curve, which can balance the precision and recall. 

As shown in Table~\ref{tab:f1:score}, the $F_1$ score of our final results are $0.887$ and $0.625$ on the validation data collected by static and dynamic platform respectively. For a static platform, the model only needs to predict the motion of moving objects. For a dynamic platform, however, the model needs to predict the motion of the robotic implicitly, which is more difficult. As a result, our method achieved better result on the static platform than on the dynamic platform. To alleviate the problem on dynamic platform, one can exploit the odometry information after perform spatial calibration and time synchronization between the LiDAR sensor and odometry. The problem is then transformed into a similar problem on the static platform. However, this may be not the ultimate way to solve this problem as the odometry information is not always accurate.

\begin{figure}
\center	
  \includegraphics[width=9cm]{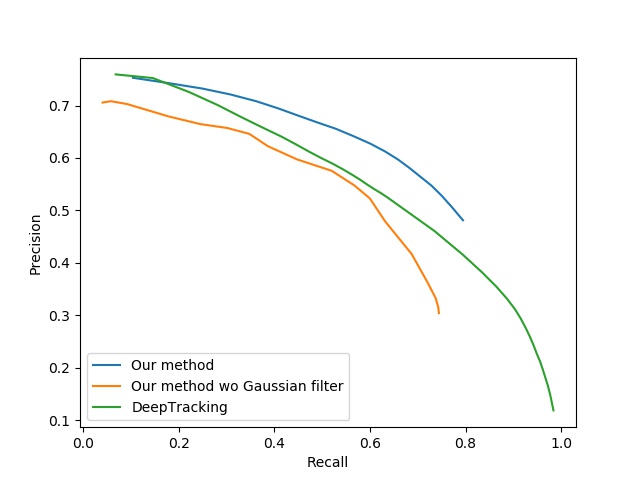}
  \caption{The precision-recall curve of occupancy map prediction results on dynamic validation data, including our full method, our method without Gaussian filter, and DeepTracking.}
  \label{fig:precision:recall}
\end{figure}

To explain the influence of Gaussian filter during training our LiDAR-FlowNet, a comparison experiment is conducted. In this experiment, we still train a LiDAR-FlowNet model but skip the Gaussian filter step and keep the other steps and parameters unchanged. The trained model is also evaluated on validation data. As shown in Table~\ref{tab:f1:score}, we can see that the $F_1$ score decreases by about $0.07$ on the dynamic platform, which demonstrates the effectiveness of Gaussian filter step in our method. At the same time, we also notice that the results are almost the same on the static platform. The reason may be that quite a number of cells remain unchanged in the map sequence from the static platform and the zero gradient problem may be negligible. The precision-recall curve of each method is also demonstrated in Fig.~\ref{fig:precision:recall}, and the same conclusion can be conducted from it.

\begin{table}
\caption{The $F_1$ score of the predicted maps obtained by our method and comparison method~\cite{DequaireIJRR2018}.}
\label{tab:f1:score}
\begin{center}
\begin{tabular}{ @{\hspace{2mm}} l @{\hspace{2mm}} c @{\hspace{2mm}} c @{\hspace{2mm}}}
\hline \hline
\multirow{2}{*}{Method}  & \multicolumn{2}{c}{$F_1$ score} \\
                         &  Static platform & Dynamic platform \\
\hline
DeepTracking~\cite{DequaireIJRR2018}  &  $0.862$ & $0.570$ \\ 
Our method w/o Gaussian filter  &  $0.884$ & $0.556$ \\ 
Our final method  &  ${\bf0.887}$ & ${\bf0.625}$ \\ 
\hline \hline
\end{tabular}
\end{center}
\end{table}

\begin{figure*}
\center
  \includegraphics[width=17.5cm]{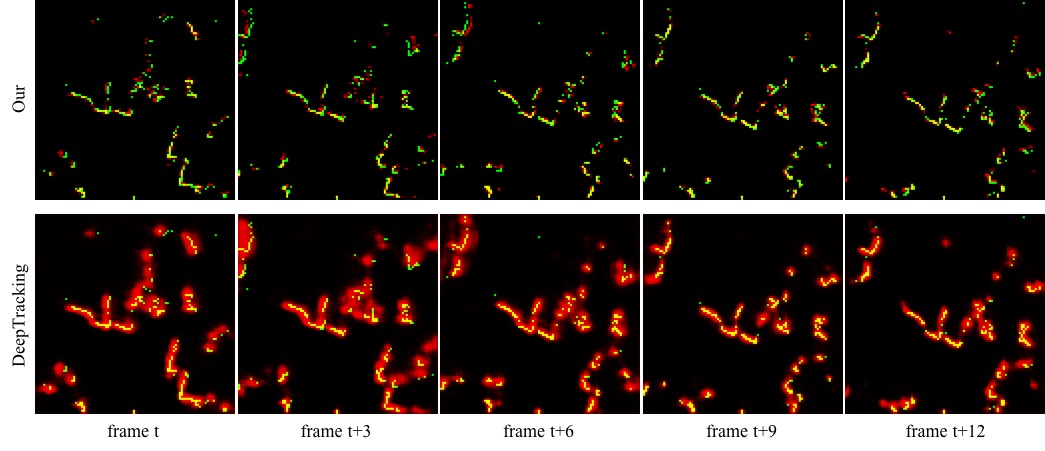}
  \caption{Some occupancy maps prediction results of our method and DeepTracking. We can see that DeepTracking can predict some occluded objects. However, it also obtains many false positives. With the help of motion flow, our method can alleviate this problem.}
  \label{fig:dmp:result}
\end{figure*}

\begin{figure*}
\center
  \includegraphics[width=16.6cm]{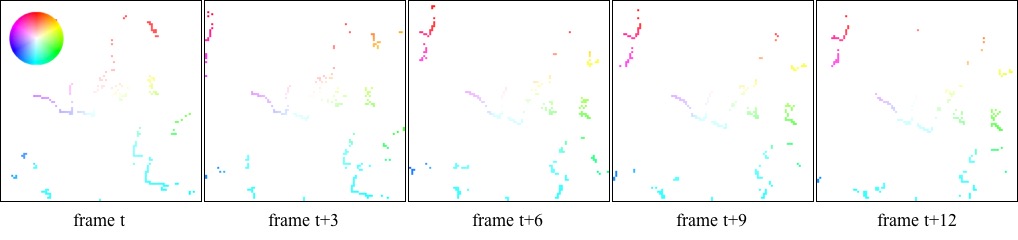}
  \caption{Some motion flow results of our method, which demonstrates that our model actually can estimate the motion of each cell.}
  \label{fig:flow}
\end{figure*}

We also compare our method with the DeepTracking~\cite{DequaireIJRR2018}, which is among the state of the arts. To be fair, we repeat DeepTracking and use the same hyperparameters to train the DeepTracking model on our training data and evaluate the model on our validation data. Unlike our method, DeepTracking directly predicts the LiDAR map. We also demonstrate the $F_1$ score in Table~\ref{tab:f1:score}. We can see that our method achieves a higher $F_1$ score than DeepTracking. Furthermore, we also draw the precision-recall curves in Fig.~\ref{fig:precision:recall}, which shows the same conclusion.

At last, we also visualize some results of LiDAR map predicted by our method and DeepTracking~\cite{DequaireIJRR2018}. To visualize the results, we put the ground truth LiDAR map in the green channel of the visualization image and put the predicted result in the red channel. Then if a cell is yellow, this cell is correctly predicted. As shown in Fig.~\ref{fig:dmp:result}, DeepTracking can predict some occluded objects, it also obtains many false positives. With the help of motion flow, our method can alleviate this problem.

\subsection{Motion Flow Estimation Results}

In this section, we illustrate the predicted motion flow as it plays a key role in our method. As shown in Fig.~\ref{fig:flow}, we visualize some motion flow results. We can see that our model actually can estimate the motion flow of each cell. However, a quantitative evaluation is still an open problem as it is difficult to collect 2D LiDAR dataset with motion flow ground truth. We leave this problem as future work. Moreover, inspired by the synthetic dataset, \eg~FlyingChairs~\cite{Dosovitskiy2015}, it is also an alternative solution to synthesize data along with ground truth. 

\section{Discussion and Conclusion}

In this paper, we propose a method to predict the 2D LiDAR map at the next moment using motion flow. This problem is challenging due to the featureless 2D LiDAR maps. To alleviate this challenge, we propose to estimate the motion flow of 2D LiDAR via the powerful deep neural networks inspired by its successful application in estimating the optical flow of the video. To this end, we design a recurrent network based on gated recurrent unit, which is named LiDAR-FlowNet. Our LiDAR-FlowNet can simultaneously estimate forward and backward motion flow between the current frame and the next frame only from the current frame and past frames. A self-supervised strategy is further designed to train the LiDAR-FlowNet model effectively. No training data need to be annotated with this strategy. With the bidirectional motion flow, it is straightforward to perform some perception tasks, \eg~with the backward motion flow, we can predict the next frame. Experimental results verify the effectiveness of our LiDAR-FlowNet as well as the proposed training strategy. Moreover, the estimated motion flow can also be used to perform other tasks, \eg~with the forward motion flow, we can detect the moving objects and separate them from the static background. We leave applications like this as our future work.

\section*{ACKNOWLEDGMENT}

This work was supported by grants from the National Basic Research Program of China (2015CB351806), the National Natural Science Foundation of China (61825101) and China Postdoctoral Science Foundation.




\bibliographystyle{IEEEtran}
\bibliography{mod.bib}

\end{document}